\documentclass{article}
\usepackage{spconf,amsmath}
\usepackage{graphicx}
\usepackage{here}
\usepackage{paralist}
\usepackage{multirow}
\usepackage{color, soul}
\usepackage{booktabs}
\usepackage{seqsplit}
\usepackage{multirow}
\usepackage {tablefootnote}
\usepackage[hidelinks]{hyperref}
\usepackage{cite}

\title{Exploration of Language Dependency\\for Japanese Self-Supervised Speech Representation Models}
\name{Takanori Ashihara, Takafumi Moriya, Kohei Matsuura, Tomohiro Tanaka}
\address{NTT Corporation, Japan} 

\begin{document}
\ninept
\maketitle
\begin{abstract}
Self-supervised learning (SSL) has been dramatically successful not only in monolingual but also in cross-lingual settings.
However, since the two settings have been studied individually in general, there has been little research focusing on how effective a cross-lingual model is in comparison with a monolingual model.
In this paper, we investigate this fundamental question empirically with Japanese automatic speech recognition (ASR) tasks.
First, we begin by comparing the ASR performance of cross-lingual and monolingual models for two different language tasks while keeping the acoustic domain as identical as possible.
Then, we examine how much unlabeled data collected in Japanese is needed to achieve performance comparable to a cross-lingual model pre-trained with tens of thousands of hours of English and/or multilingual data.
Finally, we extensively investigate the effectiveness of SSL in Japanese and demonstrate state-of-the-art performance on multiple ASR tasks.
Since there is no comprehensive SSL study for Japanese, we hope this study will guide Japanese SSL research.
\end{abstract}


\vspace{-0.2cm}
\section{Introduction}
\vspace{-0.2cm}
\label{sec:intro}
In recent years, self-supervised learning (SSL) has attracted much attention in the speech community~\cite{ssl_review}.
Since SSL is unsupervised representation learning, pre-trained models can capture speech information contained in unlabeled data.
A variety of SSL methods have been proposed and feature impressive performance, especially in the automatic speech recognition (ASR) task in English~\cite{cpc, w2v2, hubert, wavlm, data2vec, w2v_bert}.
Subsequent studies have also shown task generalizability with better performance than hand-crafted features (e.g.,~log mel filterbank output) for a variety of downstream tasks such as speaker verification and emotion recognition~\cite{superb, ashihara}.
\par
\begin{figure}[t]
  \centering
  \includegraphics[width=8.5cm]{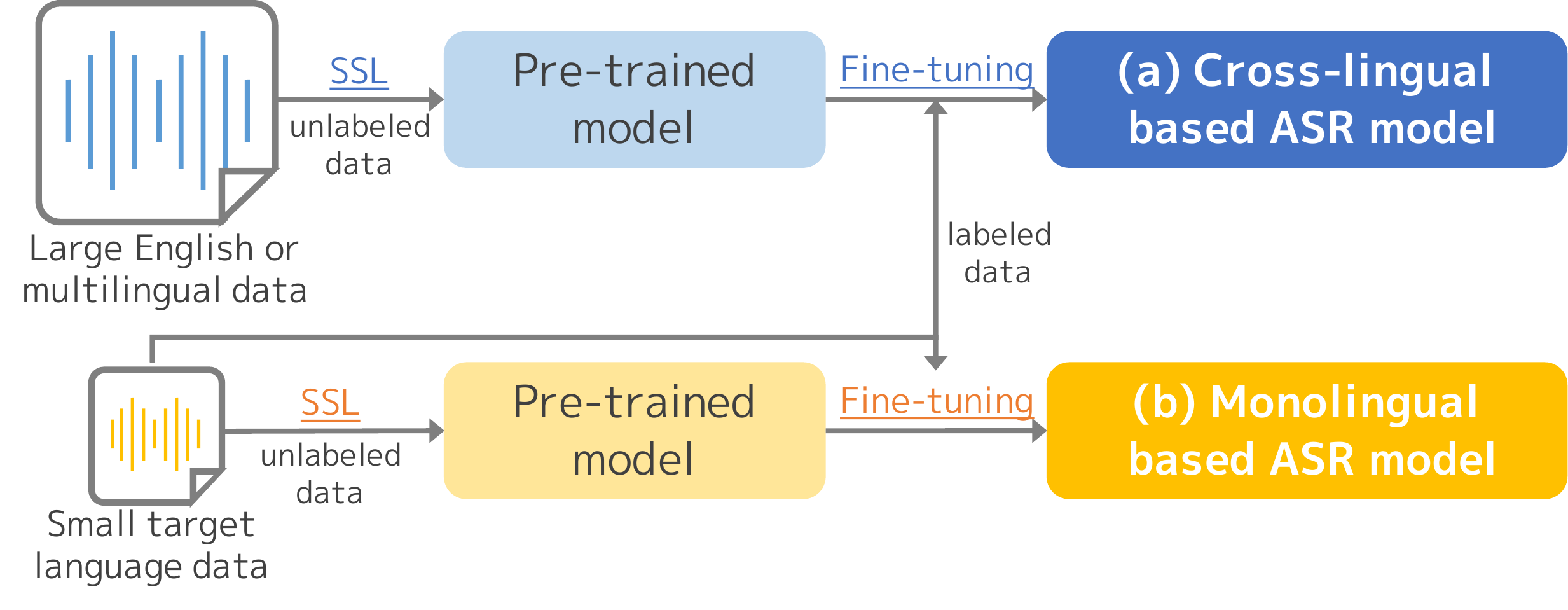}
  \vspace{-0.3cm}
  \caption{SSL-based models for target languages.}
  \label{fig:intro}
  \vspace{-0.5cm}
\end{figure}

In addition to task generalizability, SSL models have demonstrated language generalizability to alleviate the problem of data scarcity.
For example, some studies have built a multilingual or English pre-trained model for improving performance not only for English but also for low-resource languages~\cite{clossl1, kawakami, xlsr, xlsr_large}.
This universal representation democratizes speech tasks in a low-resource language, and thereby, various models in desired languages are publicly available from open-access platforms such as the Hugging~Face model hub\footnote{\url{https://huggingface.co/models}}.
In the above studies and released models, to evaluate transferability to other languages, the downstream non-English models are typically constructed in the context of transfer learning, in which a multilingual or English model is fine-tuned with a small amount of target language data (model (a) shown in Figure~\ref{fig:intro}).
\par
However, despite the success of cross-lingual models, there has been little understanding of how effective the language transferability of multilingual models is in comparison with monolingual models pre-trained and fine-tuned in a monolingual setting (model (b) in Figure~\ref{fig:intro}).
In other words, how strongly does the SSL model depend on the language?
More concretely, we can ask if training data can be collected for the target language sufficiently, can a monolingual model achieve higher accuracy than a cross-lingual model?
And if so, how much unlabeled data in the target language is required for pre-training to achieve higher accuracy than a multilingual/English model that is pre-trained with tens of thousands of hours?
Due to the substantial computing and energy costs of attempting SSL many times, this fundamental question is critical to determining whether pre-training is expected to achieve sufficient performance.
For example, some low-resource domains cannot reap the benefits of generalizability~\cite{violeta}, and hence, this implication would help to identify possible applications of SSL.
Furthermore, in practical scenarios, it is often possible to collect hundreds of hours of non-English unlabeled data for languages that have a certain number of daily speakers, such as Asian or Indian languages.
\par
In this paper, we focus on Japanese SSL with open-access corpora.
First, we compare the ASR performance of models (a) and (b) illustrated in Figure~\ref{fig:intro} to identify the language dependency of SSL on the JNAS dataset~\cite{jnas}.
Since the JNAS is a low-resource corpus containing a set of newspaper readings in Japanese, the acoustical domain is similar to CommonVoice~\cite{commonvoice}, LibriSpeech~\cite{librispeech}, Multilingual LibriSpeech~\cite{mls}, and Libri-Light~\cite{librilight} except for the language.
Second, we analyze the relationship between the amount of Japanese unlabeled data and the performance to find out how much data is needed to rank with the multilingual SSL model on the JNAS task.
The results of the above exploration demonstrate that language dependency is crucial rather than acoustical dependency, and furthermore, a monolingual model pre-trained on 200 hours shows comparable performance to a multilingual model.
Finally, to the best of our knowledge, since there are no articles on Japanese SSL unlike other non-English languages~\cite{china, swedish, ind}, we extensively investigate an SSL-based ASR model and achieve state-of-the-art error rates on multiple Japanese ASR tasks.\footnote{Some pre-trained models in Japanese will be released to accelerate Japanese SSL research at the following URL: \url{https://huggingface.co/ashi-ta/japanese-pretrained-ckpts}}

\vspace{-0.3cm}
\section{Related work}
\vspace{-0.3cm}
Previous studies have reported on the impact of dataset domains for SSL.
For example,~\cite{robust} investigated how the in-domain and out-of-domain datasets used for pre-training affect the downstream ASR performance and demonstrated that an in-domain dataset is effective in any situation.
\cite{fast} also explored the effect of dataset bias such as the gender or speed of speech.
These works mainly focused on how much the dataset domain of unlabeled data affects the performance of downstream tasks in English, whereas not much was elucidated about language transferability/dependency.
Because SSL representations have been suggested to be partly correlated phonemic structures that have a certain relationship to language~\cite{cpc}, this paper investigates how much language dependency affects the performance in a downstream ASR system.

\vspace{-0.2cm}
\section{Method}
\vspace{-0.2cm}
\label{sec:method}
\subsection{Self-supervised learning}
\vspace{-0.1cm}
\label{ssec:ssl}
\subsubsection{wav2vec 2.0}
\label{sssec:w2v2}
In this paper, we focus on wav2vec 2.0~\cite{w2v2} among the various SSL methods.
In general, the architecture of a pre-trained model consists of two main components:
\begin{gather}
\mathbf{H}^{0} = \mathrm{Convolution}(\mathbf{X}),\\
\mathbf{H}^{l} = \mathrm{Encoder}(\mathbf{H}^{l-1}),
\end{gather}
where $\mathrm{Convolution}(\cdot)$ is a multilayer convolutional network (CNN) that maps input audio $\mathbf{X}$ to latent feature $\mathbf{Z} = \{\mathbf{z}_1, \cdots, \mathbf{z}_T\}$ with downsampling.
After the first components, the representation is fed into an $\mathrm{Encoder}(\cdot)$ model such as Transformer to output hidden speech representation $\mathbf{H}^{l} = \{\mathbf{h}_1^{l}, \cdots, \mathbf{h}_T^{l}\}$.
Here, $l$ and $T$ are the block index of $\mathrm{Encoder}(\cdot)$ and output sequence length, respectively.
We define the final output of $\mathrm{Encoder}$ $\mathbf{H}^{l}$ as context representations $\mathbf{C} = \{\mathbf{c}_1, \cdots, \mathbf{c}_T\}$.
\par
For training objectives, wav2vec 2.0 utilize the quantized representation $\mathbf{Q}=\{\mathbf{q}_1, \cdots, \mathbf{q}_T\}$ calculated from $\mathbf{Z}$ as a training target.
When pre-training, the input to $\mathrm{Encoder}(\cdot)$ is masked at a certain proportion of time steps, and the contrastive loss is computed from $\mathbf{C}$ and $\mathbf{Q}$.

\vspace{-0.1cm}
\subsubsection{Input audio feature}
\vspace{-0.1cm}
\label{sssec:inputtype}
Although the input to $\mathrm{Convolution}(\cdot)$ generally uses a raw waveform (WAV), deep CNNs may not be able to acquire an adequate feature representation from a small amount of WAVs.
Therefore, we investigate not only a WAV but also log mel-filterbank output (FBANK) for inputting to SSL models~\cite{push}.
Specifically, we extract 80-dimensional FBANK features with a 25 ms window width and 10 ms shift extracted from raw audio.
To normalize the features, we subject the data set to pre-estimated global cepstral mean and variance normalization (CMVN).
With the application of FBANK, the composition of $\mathrm{Convolution}(\cdot)$ is also modified from the 7-L 1-D CNNs used in the original wav2vec 2.0 model to the 2-L 2-D convolutional layers.
Each 2-D convolution layer has 32 channels with a kernel size of 3 and stride of 2, and hence the output temporal resolution is reduced by a factor of 4 against the FBANK sequence, resulting in the reduction of GPU memory consumption.

\vspace{-0.1cm}
\subsubsection{Encoder architecture}
\vspace{-0.2cm}
\label{sssec:encoder}
In addition to the type of audio input, we compare the Conformer~\cite{conformer} encoder with the Transformer encoder for the $\mathrm{Encoder}(\cdot)$ component.
Recent studies have reported the effectiveness of the Conformer encoder for SSL in English~\cite{push, w2vaug}, but there is no comparative study in Japanese.
In this paper, we adopt a Conformer block comprised of a multi-head self-attention module with relative position encoding, a convolution module, and a feed-forward module.

\vspace{-0.2cm}
\subsection{ASR system}
\vspace{-0.2cm}
\label{ssec:asr}
After SSL, the pre-trained model is optimized using a Connectionist Temporal Classification (CTC)~\cite{ctc} criterion.
When fine-tuning, we investigate where to apply feature masking similar to SpecAugment~\cite{specaug}.
In the original wav2vec 2.0 paper, this operation is applied after $\mathrm{Convolution}(\cdot)$ (post-CNN) as temporal/channel masking.
However, when inputting FBANK, the same transformation can be applied before the $\mathrm{Convolution}(\cdot)$ (pre-CNN) as temporal/spectral masking which is more similar to the original SpecAugment.
Because the convolutional module can also be regularized in addition to $\mathrm{Encoder}(\cdot)$, the ASR performance can be further improved for the FBANK-based model.

\vspace{-0.2cm}
\section{Experimental setup}
\vspace{-0.2cm}
\label{sec:expset}
\subsection{Dataset}
\vspace{-0.3cm}
\label{subsec:data}
For Japanese speech data, we utilized four publicly available corpora: the Corpus of Spontaneous Japanese (\texttt{CSJ})~\cite{csj}, \seqsplit{LaboroTVSpeech} (\texttt{LTVS})~\cite{laboro}, Japanese Newspaper Article Sentences (\texttt{JNAS})~\cite{jnas}, and the Japanese speech corpus of Saruwatari Laboratory, the University of Tokyo (\texttt{JSUT})~\cite{jsut}.
Table~\ref{tab:data} summarizes the domain and the amount of datasets after preparation with each recipe in ESPnet~\cite{espnet}.
Since \texttt{JNAS} is recorded from speakers reading newspaper articles, the acoustical domain is close to LibriSpeech (\texttt{LS})~\cite{librispeech}, CommonVoice~\cite{commonvoice}, Multilingual LibriSpeech~\cite{mls} and Libri-Light~\cite{librilight}, which are derived from audiobooks or reading sentences.
On the other hand, \texttt{CSJ} contains a collection of academic presentation speeches and simulated public speeches, which is analogous to the TED-LIUM-v2 (\texttt{TED2}) corpus~\cite{ted2} collected from TED talks.
\par
For SSL, as unlabeled data, we utilized only high-resource corpora, specifically \texttt{CSJ}, \texttt{LTVS}, and \texttt{LS}.
For fine-tuning, we defined six ASR tasks with full labeled data and categorized them into two resource-based situations; \texttt{CSJ}, \texttt{LTVS}, and \texttt{LS} belong to the high-resource situation, and \texttt{JNAS}, \texttt{JSUT}, and \texttt{TED50h} belong to the low-resource one.
Here, \texttt{TED50h} is an English dataset randomly extracted from the \texttt{TED2} corpus for only 50 hours to make the amount of data roughly equivalent to \texttt{JNAS}.
We utilized all data contained in each corpus unless otherwise stated.

\begin{table}[t]
\caption{Domain and amount of Japanese datasets in hours after pre-processing by using ESPnet recipes. Each column shows amount of data for training, development, and multiple evaluation sets. Considering amount of training data, four tasks can be divided into top two (high-resource) and bottom two (low-resource) rows.}
\vspace{-0.3cm}
\label{tab:data}
\begin{center}
\scalebox{0.88} [0.88] {
\begin{tabular}{c|c|ccc}
\toprule
Dataset               & Domain      & Train                   & Dev                  & Test                       \\
\midrule
\multirow{2}{*}{\texttt{CSJ}}  & Spontaneous & \multirow{2}{*}{515.6}  & \multirow{2}{*}{6.5} & dev / eval1 / eval2 / eval3      \\ 
                      & speech      &                         &                      & 6.5 / 1.8 / 1.9 / 1.3            \\
\midrule
\multirow{2}{*}{\texttt{LTVS}} & Broadcast   & \multirow{2}{*}{2031.2} & \multirow{2}{*}{5.0} & dev / dev-4k / tedx-jp-10k \\  
                      & speech      &                         &                      & 5.0 / 13.7 / 8.9           \\
\midrule
\midrule
\multirow{2}{*}{\texttt{JNAS}} & Newspaper   & \multirow{2}{*}{47.8}   & \multirow{2}{*}{2.5} & dev / testset-100 / testset-500  \\   
                      & readings    &                         &                      & 2.5 / 1.6 / 1.9                  \\
\midrule
\multirow{2}{*}{\texttt{JSUT}} & Text        & \multirow{2}{*}{9.7}    & \multirow{2}{*}{0.3} & dev / eval1                \\   
                      & readings    &                         &                      & 0.3 / 0.3                 \\
\bottomrule
\end{tabular}
}
\end{center}
\vspace{-0.3cm}
\end{table}

\vspace{-0.2cm}
\subsection{Training and evaluation}
\vspace{-0.2cm}
\label{subsec:train_and_eval}
\subsubsection{Pre-training}
\vspace{-0.2cm}
The model architecture explored in this paper is based on~\cite{w2v2} and we used the same setting as the wav2vec 2.0 {\sc Base} model for the Transformer encoder.
As the Conformer encoder, we used 12 blocks containing a multi-head self-attention with 768-dim embeddings and 12 heads, a convolution module with a kernel size of 31, and a 3072-dim feed-forward module.
We basically operated with16-bit floating point (FP16) arithmetic but with FP32 only when we encountered the gradient overflow issue.
Note that the hyperparameters related to input length were converted from sample units to frame units when pre-training and fine-tuning with FBANK as input.
For example, if the maximum batch size per GPU is 87.5 seconds for WAV input, then there are 8750 frames for FBANK input.
\par
In our experiment, we also compared with released pre-trained models available from fairseq.\footnote{\url{https://github.com/facebookresearch/fairseq/tree/main/examples/wav2vec}}

\vspace{-0.2cm}
\subsubsection{Fine-tuning}
\vspace{-0.2cm}
We fine-tuned the SSL model with two types of configurations in relation to the resource-based situations described in Section~\ref{subsec:data}.
For the high-resource situation, we used the same setup as in the case where the wav2vec 2.0 {\sc Base} model was fine-tuned with \texttt{LS} 960h~\cite{w2v2} except for the parameters for masking as described below.
The other type is for the low-resource situation, with the maximum number of training steps reduced to 80 k and the peak of the learning rate reduced to $3\times10^{-5}$.
As stated in Section~\ref{ssec:asr}, during fine-tuning, we adopt two positions and hyperparameters for masking: temporal (channel) masking with a length of 10 (64) and probability of 0.5 (0.1) for the post-CNN, and temporal (spectral) masking with a length of 20 (30) and probability of 0.65 (0.1) for the pre-CNN.
In addition, we freeze the feature encoder $\mathrm{Convolution}(\cdot)$ only when applying post-CNN masking.
The output vocabulary was characters in all tasks, and the vocabulary number was 3259, 3949, 2272, 2739, 28 and 31 for the \texttt{CSJ}, \texttt{LTVS}, \texttt{JNAS}, \texttt{JSUT}, \texttt{LS} and \texttt{TED50h} tasks, respectively.

\vspace{-0.2cm}
\subsubsection{Decoding}
\vspace{-0.2cm}
We performed Viterbi decoding using a CTC model that averaged the weights of multiple checkpoints.
During fine-tuning, we evaluated the model every 6400 steps for high-resource tasks and 1600 steps for low-resource tasks and stored the top-5 checkpoints for averaging.
Here, the evaluation measure was the character error rate (CER) for Japanese tasks and word error rate (WER) for English tasks.
\par
All the above training and decoding processes were implemented with fairseq\footnote{\url{https://github.com/facebookresearch/fairseq}}~\cite{fairseq}.

\vspace{-0.2cm}
\section{Results}
\label{sec:res}
\vspace{-0.2cm}
Here, we first offer the analysis results from Section~\ref{subsec:res1} to Section~\ref{subsec:res3} and then provide the performances of our SSL-based ASR models for various Japanese tasks.
For the sake of simplicity, we used a Transformer-based SSL model fine-tuned with post-CNN masking except in Section~\ref{subsec:res4}, 

\vspace{-0.2cm}
\subsection{WAV vs. FBANK}
\vspace{-0.2cm}
\label{subsec:res1}
First of all, to verify the impact of the input audio type used in our experiment, we compared the performance of the models inputting WAV and FBANK.
Table~\ref{tab:res1} shows the results for the \texttt{LS} and \texttt{CSJ} models, which were pre-trained and fine-tuned with the same dataset.
There was no severe degradation for \texttt{LS} but rather a performance improvement for \texttt{CSJ}.
This is possible because the amount of data in \texttt{CSJ} is approximately half the volume of \texttt{LS}, leading to the difficulty of optimizing deep CNNs that encode WAV to latent features.
On the basis of these results, subsequent analyses basically use the FBANK input model for Japanese SSL.

\begin{table}[t]
\caption{Effect of input type on \texttt{LS} and \texttt{CSJ} when pre-training and fine-tuning on same data. Note that WAV input model of \texttt{LS} is publicly available wav2vec 2.0 {\sc Base}.}
\vspace{-0.35cm}
\label{tab:res1}
\begin{center}
\scalebox{0.85} [0.85] {
\begin{tabular}{cc|cccc}
\toprule
(Un)labeled          & Input & \multicolumn{4}{c}{WER}  \\   \cline{3-6}
data                 & type  & dev-clean & dev-other & test-clean & test-other \\
\midrule
\multirow{2}{*}{\texttt{LS}}  & WAV   & 3.2       & 8.9       & 3.4        & 8.4        \\
                     & FBANK & 3.2       & 9.0       &   3.4    &    8.7        \\
\midrule
\midrule
(Un)labeled          & Input & \multicolumn{4}{c}{CER}  \\ \cline{3-6}
data  & type &    dev   &   eval1   &      eval2                 &    eval3                \\
\midrule
\multirow{2}{*}{\texttt{CSJ}} & WAV   &   4.0  & 4.7                  &    3.5                  &    3.8                 \\
                     & FBANK        &   3.4       &     4.6                  &     3.2                   &     3.7                 \\
\bottomrule
\end{tabular}
}
\end{center}
\vspace{-0.5cm}
\end{table}

\vspace{-0.3cm}
\subsection{Language dependency}
\label{subsec:res2}
\begin{figure}[t]
  \centering
  \includegraphics[width=8.8cm]{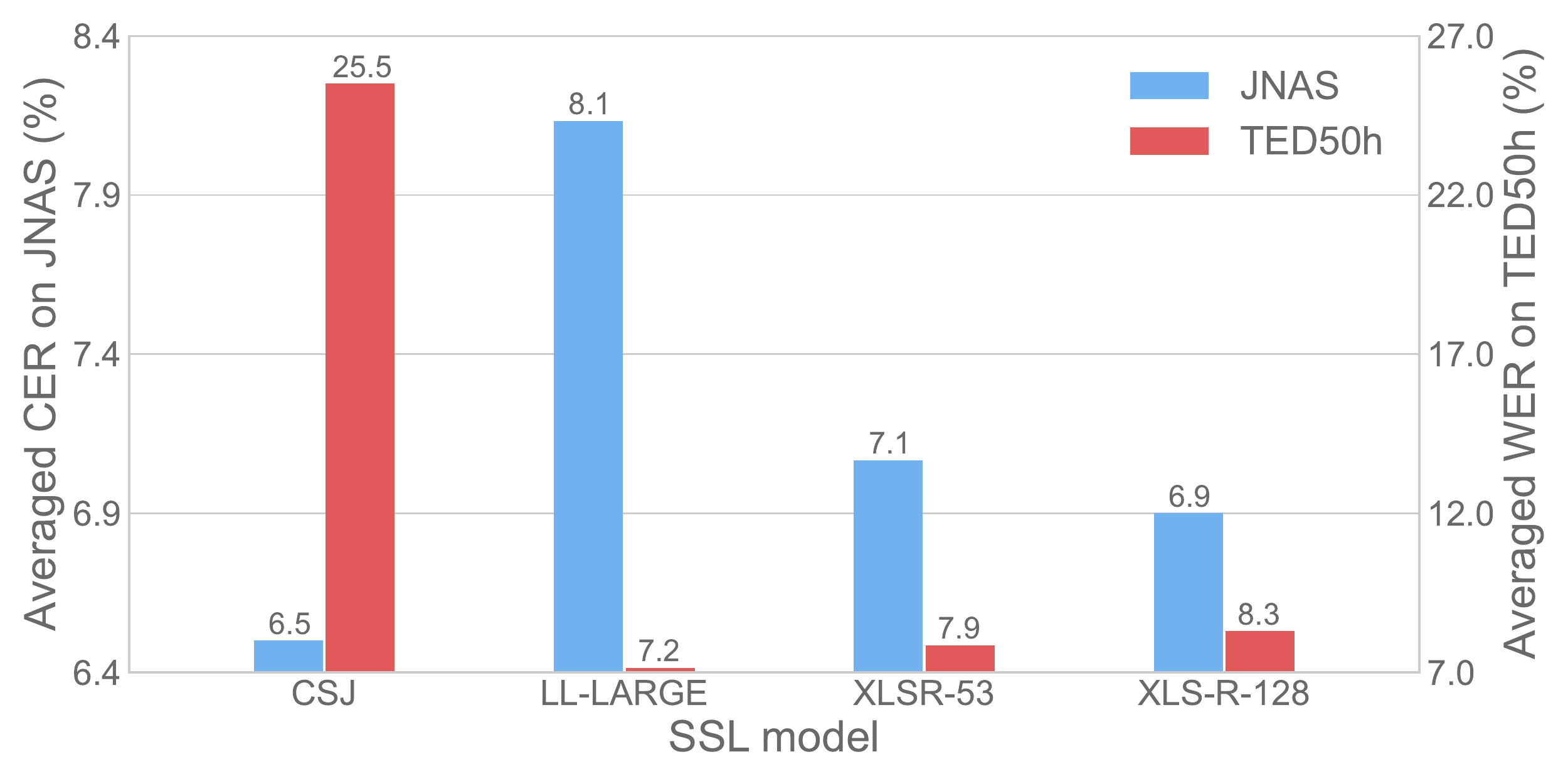}
  \vspace{-0.7cm}
  \caption{Evaluation result for different SSL models on \texttt{JNAS} (left-bar) and \texttt{TED50h} (right-bar) tasks. LL-LARGE is taken from released wav2vec 2.0 {\sc Large} model pre-trained on Libri-Light~\cite{librilight}.}
  \vspace{-0.5cm}
  \label{fig:res1}
\end{figure}

\vspace{-0.2cm}
For this section, we analyze the language dependency with four SSL models: the Japanese model pre-trained on \texttt{CSJ}, a publicly available wav2vec 2.0 {\sc Large} model pre-trained on Libri-Light~\cite{librilight}, XLSR-53~\cite{xlsr}, and XLS-R-128\footnote{We used XLS-R (0.3B) pre-trained on 128 languages with the same architecture as XLSR-53 from the following link:\\ \url{https://github.com/facebookresearch/fairseq/tree/main/examples/wav2vec/xlsr}}~\cite{xlsr_large}.
Figure~\ref{fig:res1} illustrate the performances of two low-resource ASR tasks in Japanese (\texttt{JNAS}; left-bar) and in English (\texttt{TED50h}; right-bar).
Note that each error rate is averaged over all test sets within each task.
The result shows that the multilingual models showed clear language generalizability (7.1\%/7.9\% for XLSR-53 and 6.9\%/8.3\% for XLS-R-128), but each monolingual model had lower error rates in the corresponding language task; the CSJ model achieved 6.5\%/25.5\%, and the LL-LARGE model achieved 8.1\%/7.2\% on the \texttt{JNAS}/\texttt{TED50h} task.
Because the dataset domain of the CSJ and LL-LARGE SSL models is similar to \texttt{TED50h} and \texttt{JNAS}, respectively, this indicates that it is critical to pre-train and fine-tune in the same language rather than in the same acoustical domain.

\par

\vspace{-0.2cm}
\subsection{Amount of unlabeled data}
\label{subsec:res3}
\begin{figure}[t]
  \centering
  \includegraphics[width=8.8cm]{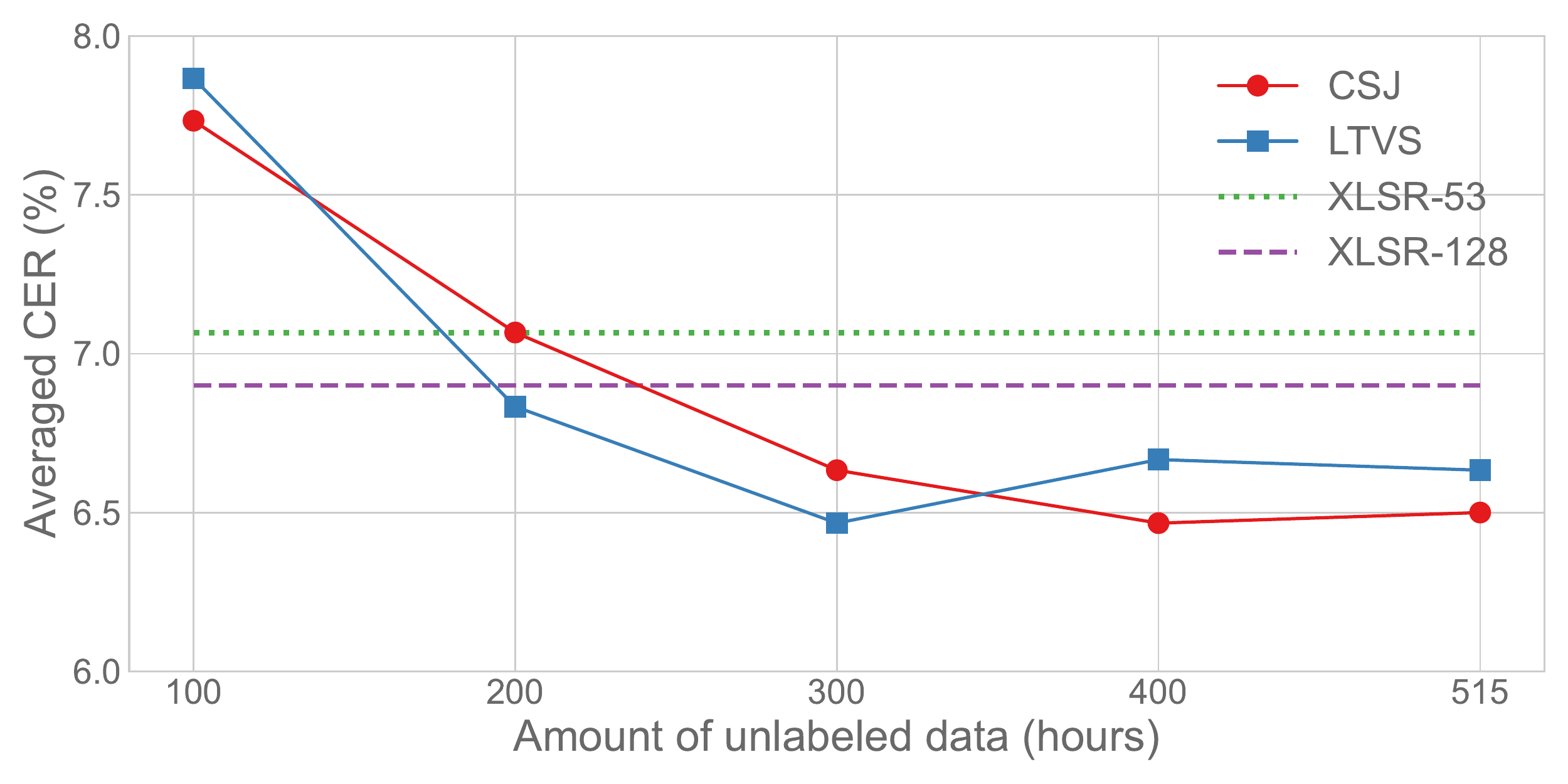}
  \vspace{-0.5cm}
  \caption{Evaluation result with different amounts of unlabeled data on \texttt{JNAS}, which is out-of-domain setting for all models.}
  \vspace{-0.2cm}
  \label{fig:res2}
\end{figure}

\vspace{-0.2cm}
Section~\ref{subsec:res2} shows results emphasizing the language dependency of SSL rather than the acoustical dependency and obvious language generalizability of the multilingual model. 
However the above results are only possible when a sufficient amount of unlabeled data in the target language has been collected such as \texttt{CSJ}.
Therefore, we assume several conditions with different amounts of unlabeled data in Japanese and investigate how much collected unlabeled data would reverse the performance of the multilingual model.
Specifically, in this analysis, we built Japanese SSL models on five different data volumes for comparison with the multilingual models and performed the same experiment with \texttt{LTVS} as well as \texttt{CSJ} for confirmation.
Figure~\ref{fig:res2} presents the relationship between the number of Japanese unlabeled data and the CERs averaged over three evaluation sets on the \texttt{JNAS} task.
The results indicated that the models pre-trained in approximately 200 hours achieved comparable performance to the multilingual models, and this trend is seemingly almost the same for both \texttt{CSJ} and \texttt{LTVS}.

\subsection{Comparison on Japanese ASR tasks}
\label{subsec:res4}
In Section~\ref{subsec:res3}, the performance of the multilingual model caught up to the monolingual models with nearly 200 hours of unlabeled data which is much smaller than the amount for only the open-access corpora in Japanese.
Therefore, we built SSL models with a combination of high-resource data (i.e.,~\texttt{CSJ} and \texttt{LTVS}) and compared various models fine-tuned with the training techniques introduced in Section~\ref{sec:method}.
Table~\ref{tab:results} shows the CERs of two baselines (B1--B2), monolingual SSL-based models (M1--M6) in Japanese, cross-lingual models (C1, C2), and the existing models (E) in the four Japanese tasks.
Since the performance trend is similar between the tasks except for the masking position configuration, we report the full results only for the \texttt{CSJ} task and the best performances for other tasks due to space limitations.
Note that we exhibit only the performance of the model fine-tuned with a post-CNN for low-resource tasks because it performed better than the pre-CNN unlike for the high-resource tasks.
From the \texttt{CSJ} results, we can see that the SSL-based models (C1, C2, M1--M6) were better than the baseline models (B1--B2), of which the Japanese SSL models (M1--M6) were particularly efficient.
To verify whether other language data improves the accuracy of the target language task, we also experimented with model (M6) pre-trained with a combination of \texttt{CSJ}, \texttt{LTVS}, and \texttt{LS}, but we could not realize any additional improvement in any of the tasks.
To the best of our knowledge, the CERs of our models advance the state-of-the-art performance of the existing reports (E) in most test sets.

\begin{table}[!t]
\caption{Evaluation results with different unlabeled data, encoder, and masking strategy for \texttt{CSJ}, \texttt{LTVS}, \texttt{JNAS}, and \texttt{JSUT}. Note that only labeled data within each task is used for fine-tuning in each task (e.g.,~515.6 hours of labeled data in \texttt{CSJ} task). T and C in fourth column mean Transformer and Conformer encoder, and pre and post in third column mean pre-CNN and post-CNN masking.}
\vspace{-0.4cm}
\label{tab:results}
\begin{center}
\scalebox{0.88} [0.88] {
\begin{tabular}{cccc|cccc}
\toprule
\multicolumn{8}{c}{\texttt{CSJ}} \\
\midrule
\multirow{2}{*}{ID} & Unlabeled & \multirow{2}{*}{Enc.}    & \multirow{2}{*}{Mask} & \multicolumn{4}{c}{CER} \\  \cline{5-8}
                    & data           &                        &                     &    dev   & eval1 & eval2 & eval3 \\
\midrule
B1 & -            & T   & pre  &  4.2  & 5.3   & 3.8   & 4.1   \\
B2 & -            & C    & pre &  4.1  & 5.0   & 3.8   & 3.9   \\
\midrule
M1 & \texttt{CSJ}            & T  & post   &  3.4  & 4.6   & 3.2   & 3.7   \\
M2 & \texttt{CSJ+LTVS}    & T  & post   &  3.5  & 4.5   & 3.2   & 3.7   \\
M3 & \texttt{CSJ+LTVS}     & T  & pre   &   3.5  & 4.2   & 3.3   & 3.7   \\
M4 & \texttt{CSJ+LTVS}     & C  & post   &  3.3  & 4.2   & 3.0   & 3.5   \\
M5 & \texttt{CSJ+LTVS}    & C   & pre   &   \textbf{3.1}  & \textbf{3.9}   & \textbf{2.9}   & \textbf{3.2}   \\
M6 & \texttt{CSJ+LTVS+LS}    & C   & pre   &   3.4  & 4.3   & 3.1   & 3.4  \\
\midrule
C1 &   \multicolumn{2}{l}{XLSR-53~\cite{xlsr}}  &   post & 3.8   & 4.7   & 3.6   & 3.8   \\
C2 &   \multicolumn{2}{l}{XLS-R-128~\cite{xlsr_large}}  &   post & 3.7   & 4.7   & 3.5   & 3.8   \\
\midrule
E & \multicolumn{3}{l|}{Karita et al.~\cite{karita}}     &    \textbf{3.1}  & 4.1   & 3.2   & 3.5  \\
\bottomrule
\vspace{-0.5cm}
\end{tabular}
}
\end{center}

\begin{center}
\scalebox{0.88} [0.88] {
\begin{tabular}{cccc|ccc}
\toprule
\multicolumn{7}{c}{\texttt{LTVS}} \\
\midrule
\multirow{2}{*}{ID} & Unlabeled & \multirow{2}{*}{Enc.}    & \multirow{2}{*}{Mask} & \multicolumn{3}{c}{CER} \\  \cline{5-7}
                    & data           &                        &                     & dev & dev-4k & tedx-jp-10k \\
\midrule
B1 & -            & T   & pre  & 10.0   & 8.0   & 13.4   \\
B2 & -            & C    & pre  & 9.4   & 7.7   & 13.1   \\
\midrule
M5 & \texttt{CSJ+LTVS}    & C   & pre  & \textbf{8.7}   & 7.4   & \textbf{12.6}   \\
\midrule
E & \multicolumn{3}{l|}{ESPnet\tablefootnote{\url{https://github.com/espnet/espnet/blob/master/egs2/laborotv/asr1/README.md}}}           & 9.7  & \textbf{7.0}   & 13.0  \\
\bottomrule
\vspace{-0.5cm}
\end{tabular}
}
\end{center}

\begin{center}
\scalebox{0.88} [0.88] {
\begin{tabular}{cccc|ccc}
\toprule
\multicolumn{7}{c}{\texttt{JNAS}} \\
\midrule
\multirow{2}{*}{ID} & Unlabeled & \multirow{2}{*}{Enc.}    & \multirow{2}{*}{Mask} & \multicolumn{3}{c}{CER} \\  \cline{5-7}
                    & data           &                        &           &      dev          & testset-100 & testset-500 \\
\midrule
B1 & -            & T   & pre   &    12.0  & 8.9   & 12.4     \\
B2 & -            & C    & pre    &    14.7   & 11.1   & 14.4     \\
\midrule
M4 & \texttt{CSJ+LTVS}    & C   & post  &   \textbf{5.9}     &    \textbf{3.7}   & \textbf{5.5}   \\
\midrule
E & \multicolumn{3}{l|}{ESPnet\tablefootnote{\url{https://github.com/espnet/espnet/blob/master/egs/jnas/asr1/RESULTS}}}      &   11.0     & 8.9  & 11.3   \\
\bottomrule
\vspace{-0.5cm}
\end{tabular}
}
\end{center}

\begin{center}
\scalebox{0.88} [0.88]{
\begin{tabular}{cccc|cc}
\toprule
\multicolumn{6}{c}{\texttt{JSUT}} \\
\midrule
\multirow{2}{*}{ID} & Unlabeled & \multirow{2}{*}{Enc.}    & \multirow{2}{*}{Mask} & \multicolumn{2}{c}{CER} \\  \cline{5-6}
                    & data           &                        &                     & dev & eval1 \\
\midrule
B1 & -            & T   & pre  & 30.0   & 31.5     \\
B2 & -            & C    & pre  & 33.4   & 34.9     \\
\midrule
M4 & \texttt{CSJ+LTVS}    & C   & post  & \textbf{11.7}   & \textbf{12.6}   \\
\midrule
E & \multicolumn{3}{l|}{ESPnet\tablefootnote{\url{https://github.com/espnet/espnet/blob/master/egs2/jsut/asr1/README.md}}}           & 12.0  & 13.9   \\
\bottomrule
\end{tabular}
}
\end{center}
\vspace{-0.6cm}
\end{table}

\section{Conclusion}
\label{sec:con}
\vspace{-0.3cm}
In this paper, we empirically investigated and analyzed language dependency by comparing monolingual and cross-lingual models in Japanese ASR tasks.
Our findings indicate that the performance of the SSL-based model was more affected by the language aspect rather than the acoustic aspect during pre-training and fine-tuning, and the monolingual model pre-trained with 200 hours was comparable in accuracy to the cross-lingual models.
With this observation in mind, we built various models with modifications to the original wav2vec 2.0 model, and our models achieved state-of-the-art performance on multiple Japanese ASR tasks.

\clearpage
\bibliographystyle{IEEEbib}
\bibliography{refs}

\begin{thebibliography}{10}

\bibitem{ssl_review}
A.~Mohamed, H.-y. Lee, L.~Borgholt, J.~D. Havtorn, J.~Edin, C.~Igel,
  K.~Kirchhoff, S.-W. Li, K.~Livescu, L.~Maal{\o}e, T.~N. Sainath, and
  S.~Watanabe,
\newblock ``Self-supervised speech representation learning: {A} review,''
\newblock {\em JSTSP}, 2022.

\bibitem{cpc}
A.~v.~d. Oord, Y.~Li, and O.~Vinyals,
\newblock ``Representation learning with contrastive predictive coding,''
\newblock {\em arXiv preprint arXiv:1807.03748}, 2018.

\bibitem{w2v2}
A.~Baevski, Y.~Zhou, A.~Mohamed, and M.~Auli,
\newblock ``wav2vec 2.0: {A} framework for self-supervised learning of speech
  representations,''
\newblock in {\em NeurIPS}, 2020.

\bibitem{hubert}
W.-N. Hsu, B.~Bolte, Y.-H.~H. Tsai, K.~Lakhotia, R.~Salakhutdinov, and
  A.~Mohamed,
\newblock ``{HuBERT}: {Self}-supervised speech representation learning by
  masked prediction of hidden units,''
\newblock {\em TASLP}, 2021.

\bibitem{wavlm}
S.~Chen, C.~Wang, Z.~Chen, Y.~Wu, S.~Liu, Z.~Chen, J.~Li, N.~Kanda,
  T.~Yoshioka, X.~Xiao, J.~Wu, L.~Zhou, S.~Ren, Y.~Qian, Y.~Qian, J.~Wu,
  M.~Zeng, X.~Yu, and F.~Wei,
\newblock ``{WavLM}: {Large}-scale self-supervised pre-training for full stack
  speech processing,''
\newblock {\em JSTSP}, 2022.

\bibitem{data2vec}
A.~Baevski, W.-N. Hsu, Q.~Xu, A.~Babu, J.~Gu, and M.~Auli,
\newblock ``data2vec: {A} general framework for self-supervised learning in
  speech, vision and language,''
\newblock {\em ICML}, 2022.

\bibitem{w2v_bert}
Y.-A. Chung, Y.~Zhang, W.~Han, C.-C. Chiu, J.~Qin, R.~Pang, and Y.~Wu,
\newblock ``{w2v-BERT}: {Combining} contrastive learning and masked language
  modeling for self-supervised speech pre-training,''
\newblock in {\em ASRU}, 2021.

\bibitem{superb}
S.~w.~Yang, P.-H. Chi, Y.-S. Chuang, C.-I.~J. Lai, K.~Lakhotia, Y.~Y. Lin,
  A.~T. Liu, J.~Shi, X.~Chang, G.-T. Lin, T.-H. Huang, W.-C. Tseng, K.~t.~Lee,
  D.-R. Liu, Z.~Huang, S.~Dong, S.-W. Li, S.~Watanabe, A.~Mohamed, and
  H.~y.~Lee,
\newblock ``{SUPERB: Speech Processing Universal PERformance Benchmark},''
\newblock in {\em Interspeech}, 2021.

\bibitem{ashihara}
T.~Ashihara, T.~Moriya, K.~Matsuura, and T.~Tanaka,
\newblock ``Deep versus wide: {A}n analysis of student architectures for
  task-agnostic knowledge distillation of self-supervised speech models,''
\newblock in {\em Interspeech}, 2022.

\bibitem{clossl1}
M.~Rivi{\`e}re, A.~Joulin, P.-E. Mazar{\'e}, and E.~Dupoux,
\newblock ``Unsupervised pretraining transfers well across languages,''
\newblock in {\em ICASSP}, 2020.

\bibitem{kawakami}
K.~Kawakami, L.~Wang, C.~Dyer, P.~Blunsom, and A.~v.~d. Oord,
\newblock ``Learning robust and multilingual speech representations,''
\newblock in {\em EMNLP}, 2020.

\bibitem{xlsr}
A.~Conneau, A.~Baevski, R.~Collobert, A.~Mohamed, and M.~Auli,
\newblock ``Unsupervised cross-lingual representation learning for speech
  recognition,''
\newblock in {\em Interspeech}, 2021.

\bibitem{xlsr_large}
A.~Babu, C.~Wang, A.~Tjandra, K.~Lakhotia, Q.~Xu, N.~Goyal, K.~Singh, P.~{von
  Platen}, Y.~Saraf, J.~Pino, A.~Baevski, A.~Conneau, and M.~Auli,
\newblock ``{XLS-R}: {Self}-supervised cross-lingual speech representation
  learning at scale,''
\newblock in {\em Interspeech}, 2022.

\bibitem{violeta}
L.~P. Violeta, W.~C. Huang, and T.~Toda,
\newblock ``Investigating self-supervised pretraining frameworks for
  pathological speech recognition,''
\newblock in {\em Interspeech}, 2022.

\bibitem{jnas}
K.~Itou, M.~Yamamoto, K.~Takeda, T.~Takezawa, T.~Matsuoka, T.~Kobayashi,
  K.~Shikano, and S.~Itahashi,
\newblock ``{JNAS: Japanese speech corpus for large vocabulary continuous
  speech recognition research},''
\newblock {\em JASJ}, 1999.

\bibitem{commonvoice}
R.~Ardila, M.~Branson, K.~Davis, M.~Kohler, J.~Meyer, M.~Henretty, R.~Morais,
  L.~Saunders, F.~Tyers, and G.~Weber,
\newblock ``{Common Voice}: {A} massively-multilingual speech corpus,''
\newblock in {\em LREC}, 2020.

\bibitem{librispeech}
V.~Panayotov, G.~Chen, D.~Povey, and S.~Khudanpur,
\newblock ``{LibriSpeech}: {An ASR} corpus based on public domain audio
  books,''
\newblock in {\em ICASSP}, 2015.

\bibitem{mls}
V.~Pratap, Q.~Xu, A.~Sriram, G.~Synnaeve, and R.~Collobert,
\newblock ``{MLS}: {A} large-scale multilingual dataset for speech research,''
\newblock in {\em Interspeech}, 2020.

\bibitem{librilight}
J.~{Kahn}, M.~{Rivi{\`e}re}, W.~{Zheng}, E.~{Kharitonov}, Q.~{Xu}, P.~E.
  {Mazar{\'e}}, J.~{Karadayi}, V.~{Liptchinsky}, R.~{Collobert}, C.~{Fuegen},
  T.~{Likhomanenko}, G.~{Synnaeve}, A.~{Joulin}, A.~{Mohamed}, and E.~{Dupoux},
\newblock ``{Libri-Light}: {A} benchmark for {ASR} with limited or no
  supervision,''
\newblock in {\em ICASSP}, 2020.

\bibitem{china}
K.~Deng, Z.~Yang, S.~Watanabe, Y.~Higuchi, G.~Cheng, and P.~Zhang,
\newblock ``Improving non-autoregressive end-to-end speech recognition with
  pre-trained acoustic and language models,''
\newblock in {\em ICASSP}, 2022.

\bibitem{swedish}
R.~Al-Ghezi, Y.~Getman, A.~Rouhe, R.~Hild{\'e}n, and M.~Kurimo,
\newblock ``Self-supervised end-to-end {ASR} for low resource {L2 Swedish},''
\newblock in {\em Interspeech}, 2021.

\bibitem{ind}
K.~D. N, P.~Wang, and B.~Bozza,
\newblock ``Using large self-supervised models for low-resource speech
  recognition,''
\newblock in {\em Interspeech}, 2021.

\bibitem{robust}
W.-N. Hsu, A.~Sriram, A.~Baevski, T.~Likhomanenko, Q.~Xu, V.~Pratap, J.~Kahn,
  A.~Lee, R.~Collobert, G.~Synnaeve, and M.~Auli,
\newblock ``Robust wav2vec 2.0: {Analyzing} domain shift in self-supervised
  pre-training,''
\newblock in {\em Interspeech}, 2021.

\bibitem{fast}
Y.~Meng, Y.-H. Chou, A.~T. Liu, and H.-y. Lee,
\newblock ``Don't speak too fast: {The} impact of data bias on self-supervised
  speech models,''
\newblock in {\em ICASSP}, 2022.

\bibitem{push}
Y.~Zhang, J.~Qin, D.~S. Park, W.~Han, C.-C. Chiu, R.~Pang, Q.~V. Le, and Y.~Wu,
\newblock ``Pushing the limits of semi-supervised learning for automatic speech
  recognition,''
\newblock {\em arXiv preprint arXiv.2010.10504}, 2020.

\bibitem{conformer}
A.~Gulati, J.~Qin, C.-C. Chiu, N.~Parmar, Y.~Zhang, J.~Yu, W.~Han, S.~Wang,
  Z.~Zhang, Y.~Wu, and R.~Pang,
\newblock ``Conformer: {C}onvolution-augmented {Transformer} for speech
  recognition,''
\newblock in {\em Interspeech}, 2020.

\bibitem{w2vaug}
A.~Sriram, M.~Auli, and A.~Baevski,
\newblock ``{Wav2Vec-Aug}: {Improved} self-supervised training with limited
  data,''
\newblock in {\em Interspeech}, 2022.

\bibitem{ctc}
A.~Graves, S.~Fernandez, F.~Gomez, and J.~Schmidhuber,
\newblock ``Connectionist temporal classification: {Labelling} unsegmented
  sequence data with recurrent neural nets,''
\newblock in {\em ICML}, 2006.

\bibitem{specaug}
D.~S. Park, W.~Chan, Y.~Zhang, C.-C. Chiu, B.~Zoph, E.~D. Cubuk, and Q.~V. Le,
\newblock ``{SpecAugment}: {A} simple data augmentation method for automatic
  speech recognition,''
\newblock in {\em Interspeech}, 2019.

\bibitem{csj}
K.~Maekawa,
\newblock ``Corpus of spontaneous {Japanese}: {Its} design and evaluation,''
\newblock in {\em SSPR}, 2003.

\bibitem{laboro}
S.~Ando and H.~Fujihara,
\newblock ``{Construction of a large-scale Japanese ASR corpus on TV
  recordings},''
\newblock in {\em ICASSP}, 2021.

\bibitem{jsut}
R.~Sonobe, S.~Takamichi, and H.~Saruwatari,
\newblock ``{JSUT} corpus: {Free} large-scale {Japanese} speech corpus for
  end-to-end speech synthesis,''
\newblock {\em arXiv preprint arXiv:1711.00354}, 2017.

\bibitem{espnet}
S.~Watanabe, T.~Hori, S.~Karita, T.~Hayashi, J.~Nishitoba, Y.~Unno, N.~{Enrique
  Yalta Soplin}, J.~Heymann, M.~Wiesner, N.~Chen, A.~Renduchintala, and
  T.~Ochiai,
\newblock ``{ESPnet}: {End}-to-end speech processing toolkit,''
\newblock in {\em Interspeech}, 2018.

\bibitem{ted2}
A.~Rousseau, P.~Del{\'e}glise, and Y.~Est{\`e}ve,
\newblock ``Enhancing the {TED}-{LIUM} corpus with selected data for language
  modeling and more {TED} talks,''
\newblock in {\em LREC}, 2014.

\bibitem{fairseq}
M.~Ott, S.~Edunov, A.~Baevski, A.~Fan, S.~Gross, N.~Ng, D.~Grangier, and
  M.~Auli,
\newblock ``{fairseq}: {A} fast, extensible toolkit for sequence modeling,''
\newblock in {\em NAACL-HLT}, 2019.

\bibitem{karita}
S.~Karita, Y.~Kubo, M.~A.~U. Bacchiani, and L.~Jones,
\newblock ``A comparative study on neural architectures and training methods
  for {Japanese} speech recognition,''
\newblock in {\em Interspeech}, 2021.

\end{thebibliography}

\end{document}